\documentclass[10pt,twocolumn,letterpaper]{article} 

\usepackage{avss}
\usepackage{times}
\usepackage{epsfig}
\usepackage{graphicx}
\usepackage{amsmath}
\usepackage{amssymb}
\usepackage{todonotes}
\usepackage{soul}
\usepackage{url}

\avssfinalcopy 

\def\avssPaperID{****} 

\ifavssfinal\fi

\makeatletter
\def\ps@plain{
  \let\@oddhead\@empty
  \let\@evenhead\@empty
  \let\@oddfoot\@empty
  \let\@evenfoot\@empty
}
\makeatother
\begin{document}

\title{GeoCAM: Self-Supervised Sonar–Optical Representation Learning via Geographic Overlap Supervision}

\author{
Fadi Abdeladhim Zidi$^{1}$, \;
Andrea De Salve$^{1}$, \;
Giuseppe Celeste$^{1}$, \;
Abdellah Zakaria Sellam$^{1,2}$, \; \\
Paolo Spagnolo$^{1}$, \;
Cosimo Distante$^{1,2}$
\\[0.4em]
{\small $^{1}$Institute of Applied Sciences and Intelligent Systems (ISASI), CNR, 73100 Lecce, Italy}\\
{\small $^{2}$University of Salento, 73100 Lecce, Italy}
}

\maketitle

\begin{abstract}

Cross-modal self-supervised learning for paired side-scan sonar and underwater optical imagery is a promising direction for benthic habitat understanding, but it remains underexplored because existing methods do not fully exploit the one-to-many correspondences and geographic metadata available in real opti-acoustic surveys. This limits the ability to learn representations that are both semantically meaningful and geometrically consistent from abundant unlabeled marine data. To address this problem, we present \textbf{GeoCAM}, a geography-guided framework that learns a shared embedding space for sonar tiles and their associated optical views without semantic annotations. GeoCAM achieves its goal by combining three components: (i) a Set Attention Aggregator that adaptively summarizes multiple optical observations into a single tile-level representation; (ii) a geometry-weighted InfoNCE objective that converts GPS-derived footprint overlap into soft positive supervision; and (iii) an optional cross-modal masked reconstruction branch that encourages physically grounded cross-modal alignment. Experiments on the BenthiCat SSS-CAM dataset show that GeoCAM can be effectively exploited to learn which sonar and camera samples describe the same seabed patch, improving cross-modal retrieval and alignment over strong contrastive baselines. Code is available at \url{https://github.com/FadiZidiDz/GeoCAM}

\end{abstract}


\section{Introduction}
\label{sec:introduction}

Despite covering more than 70\% of Earth's surface, the ocean floor remains one of the least understood environments on the planet. Benthic habitat maps, which describe the spatial distribution of seafloor substrates such as sand, mud, rock, \textit{Posidonia oceanica}, and corals, are essential for marine biodiversity monitoring, conservation planning, and sustainable resource management~\cite{rajani2025benthicat}. In practice, these maps are often produced from Side-Scan Sonar (SSS) surveys collected by oceanographic vessels or Autonomous Underwater Vehicles (AUVs). SSS offers wide-area coverage and performs well under poor visibility conditions~\cite{wang2019rtseg}, but it also exhibits a low signal-to-noise ratio, inter-class ambiguity between substrates with similar backscatter patterns, and a lack of color and texture cues that are useful for fine-grained interpretation. As a result, reliable benthic classification from SSS alone remains difficult, even for trained experts~\cite{burguera2020online}.

Recent deep learning methods have improved dense seafloor segmentation in SSS imagery~\cite{zhao2021dcnet} and have also explored hybrid convolutional and recurrent transformer-based architectures for the same task~\cite{yu2021recurrent}. However, these approaches remain constrained by the need for large amounts of manual annotation, which is particularly costly in marine environments because it requires extensive survey time, expert interpretation, and ground-truth validation. This label bottleneck has limited the scale and diversity of public SSS benchmarks, and it remains one of the main obstacles to progress in benthic habitat understanding.

Self-supervised learning offers a natural way to reduce this dependence on manual labels. Contrastive learning and related representation learning methods have shown that semantically useful embeddings can be learned without annotation~\cite{chen2020simclr}. In the marine setting, this becomes particularly attractive because opti-acoustic surveys—which combine SSS with co-registered underwater cameras—provide a complementary modality: optical imagery captures color, texture, and fine-scale morphology that are often absent from SSS, while remaining georeferenced through the same survey metadata. The central challenge is therefore not the lack of signal, but how to exploit the signal already encoded in survey geometry. In this context, the BenthiCat (SSS-CAM) benchmark provides a promising yet challenging resource consisting of SSS tiles geo-referenced with matched optical imagery.
Although the BenthiCat dataset includes semantic annotations and segmentation masks for seabed classes, avoiding the use of such labels offers an opportunity to investigate learning strategies that reduce dependence on manual annotation, motivating the development of self-supervised cross-modal approaches.

However, existing cross-modal methods do not fully exploit the structure of opti-acoustic survey data. A single sonar tile is often associated with multiple optical images captured from different viewpoints and under different visibility conditions, yet most methods treat these views uniformly. In addition, the GPS-derived overlap between sonar and optical footprints provides a continuous measure of correspondence strength, but standard contrastive losses reduce this signal to a binary positive or negative label. This discards useful geometric information that could guide learning more precisely and leaves a gap between the structure available in BenthiCat SSS-CAM
and the structure exploited by current representation learning methods~\cite{radford2021clip}.

To address this problem, we introduce \textbf{GeoCAM}, a geography-guided cross-modal self-supervised framework that learns a shared embedding space for SSS tiles and their associated optical views without semantic annotations. GeoCAM is designed for the one-to-many structure of opti-acoustic surveys and combines two core components: a Set Attention Aggregator (SAA) that adaptively summarizes multiple optical observations into a single tile-level representation and a Geometry-Weighted InfoNCE objective that converts GPS-derived footprint overlap into soft positive supervision. An optional cross-modal masked reconstruction branch (CMMR) can further be enabled during pre-training to encourage physically grounded cross-modal alignment. Together, these components allow GeoCAM to learn which sonar and camera samples describe the same seabed patch while preserving the geometric consistency present in the survey itself.

The main contributions of this paper are threefold:
\begin{enumerate}
    \item We propose GeoCAM, a self-supervised framework for opti-acoustic representation learning that aligns sonar tiles and their matched optical views in a shared embedding space without requiring semantic labels.
    \item We introduce a set-aware aggregation strategy and a geometry-weighted contrastive objective that explicitly model the one-to-many structure and overlap information present in real opti-acoustic surveys.
    \item We provide a reproducible evaluation protocol on the BenthiCat SSS-CAM benchmark, establishing a strong baseline for future work on unlabeled cross-modal benthic habitat understanding.
\end{enumerate}

\section{Proposed Method}
\label{sec:method}

\begin{figure*}[t]
\centering
\includegraphics[width=\textwidth]{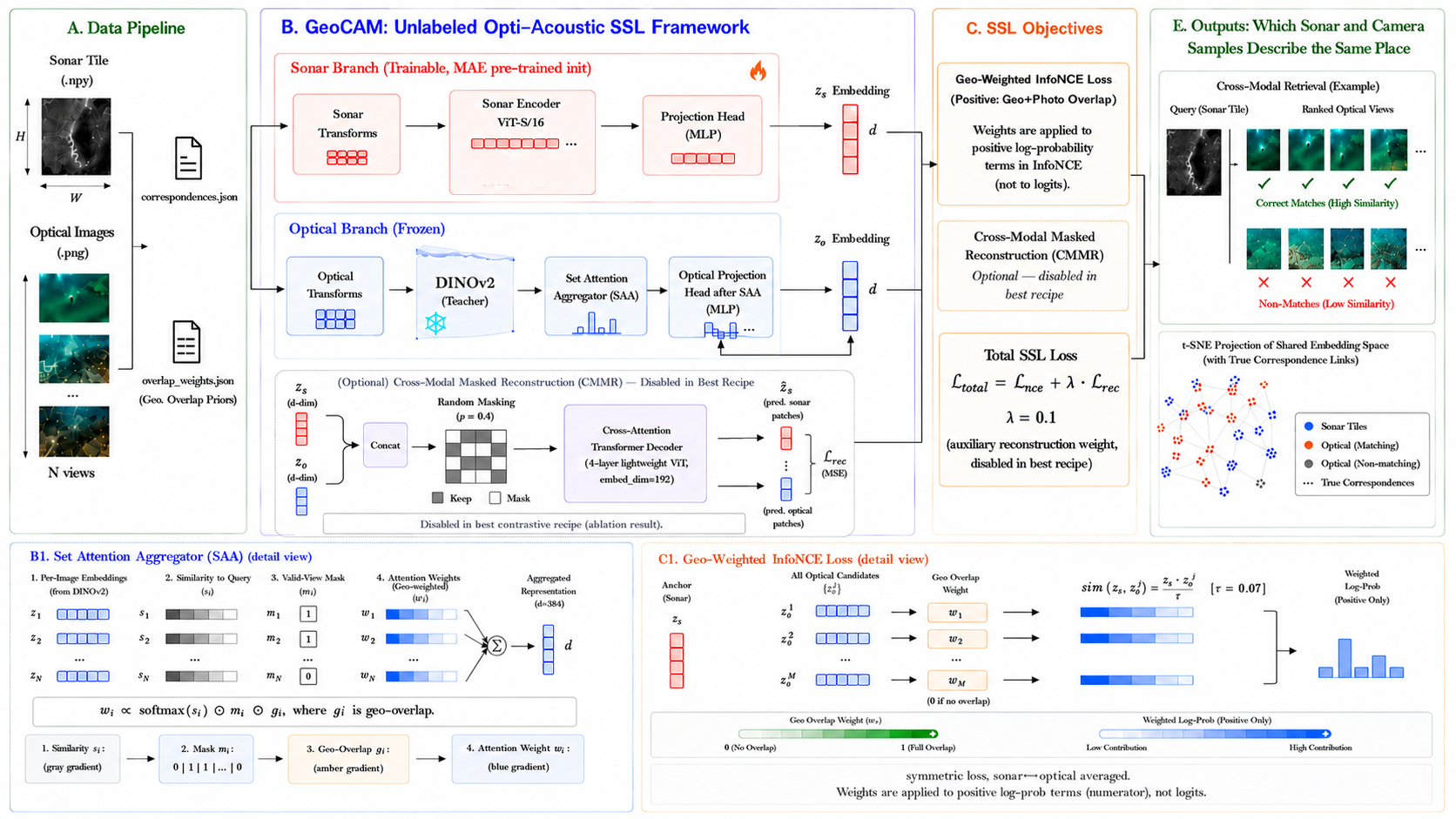}
\caption{Overview of GeoCAM, which learns a shared sonar-optical representation from unlabeled opti-acoustic correspondences (links from BenthiCat; overlap ratios computed offline via footprint polygon intersection) using set-level aggregation and geometry-weighted contrastive learning. An optional cross-modal masked reconstruction head (CMMR) can be enabled during pre-training.}
\label{fig:architecture}
\end{figure*}

GeoCAM is a geography-guided cross-modal self-supervised framework designed to align side-scan sonar (SSS) tiles and their matched underwater optical views in a shared embedding space without semantic labels. As illustrated in Figure~\ref{fig:architecture}, the method follows an asymmetric design: the sonar branch learns acoustic representations directly from sonar backscatter, while the optical branch aggregates multiple matched views into a single tile-level descriptor. The two modalities are aligned through a geometry-weighted contrastive objective that exploits footprint overlap information available in standard survey metadata.

\subsection{Architecture Overview}

For each training sample, GeoCAM receives one sonar tile and a variable-size set of matched optical images. Let the sonar input be
\begin{equation}
s \in \mathbb{R}^{1 \times H \times W},
\label{eq:sonar_input}
\end{equation}
and let the matched optical set be
\begin{equation}
\mathcal{O} = \{o_i\}_{i=1}^{N}, \qquad o_i \in \mathbb{R}^{3 \times 224 \times 224},
\label{eq:optical_set}
\end{equation}
where $N$ may vary across samples.

Each optical view is associated with an overlap prior
\begin{equation}
w_i \in [0,1],
\label{eq:overlap_prior}
\end{equation}
computed from the geographic intersection between sonar and camera footprints. These overlap ratios provide soft geometric supervision that reflects the strength of correspondence between modalities.

The sonar tile is encoded by a ViT-S/16 backbone, producing
\begin{equation}
h_s = f_s(s),
\label{eq:sonar_encoder}
\end{equation}
while each optical image is encoded by a frozen DINOv2 ViT-S/14 backbone,
\begin{equation}
h_{o_i} = f_o(o_i), \qquad i=1,\dots,N.
\label{eq:optical_encoder}
\end{equation}

Since the optical observations are unordered and may vary significantly in quality, GeoCAM aggregates the set $\{h_{o_i}\}$ into a single optical descriptor using a Set Attention Aggregator (SAA). Finally, both modalities are mapped into a shared normalized embedding space using lightweight projection heads:
\begin{equation}
\begin{aligned}
z_s &= g_s(h_s), \\
z_o &= g_o(h_o), \\
\|z_s\|_2 &= 1, \qquad \|z_o\|_2 = 1,
\end{aligned}
\label{eq:projected_embeddings}
\end{equation}
where $z_s$ and $z_o$ are used for contrastive alignment.

\subsection{Sonar Branch}

The sonar branch is responsible for learning acoustic representations that capture local seabed structure from SSS tiles. We employ a ViT-S/16 \cite{dosovitskiy2021vit} encoder as the sonar backbone and fine-tune it during training to adapt to sonar-specific texture statistics. The encoder output is projected into the shared embedding space through a small multilayer perceptron.

\subsection{Optical Branch and Set Attention Aggregation}

The optical branch processes the matched underwater images associated with each sonar tile. In opti-acoustic surveys, a single sonar tile often corresponds to multiple optical views captured under different viewpoints, illumination, and turbidity conditions. Mean pooling treats all views equally and is therefore sensitive to noisy or weakly informative frames. To address this, GeoCAM uses the Set Attention Aggregator (SAA) to learn view importance and produce a robust tile-level optical representation.

Given optical embeddings $\{h_{o_i}\}_{i=1}^N$, SAA computes attention weights
\begin{equation}
a_i = \frac{\exp\left(q^\top h_{o_i}/\sqrt{d} + \log \tilde{w}_i\right)}{\sum_{j=1}^N \exp\left(q^\top h_{o_j}/\sqrt{d} + \log \tilde{w}_j\right)},
\end{equation}
where $q$ is a learned query vector, $d$ is the embedding dimension, and $\tilde{w}_i$ is the geographic overlap weight of view $i$ ($\tilde{w}_i{=}1$ when geometry weighting is disabled). The aggregated optical representation is then computed as
\begin{equation}
h_o = \sum_{i=1}^N a_i h_{o_i}.
\end{equation}

This permutation-invariant design preserves the one-to-many correspondence structure while downweighting visually degraded views such as motion-blurred or highly turbid images.

\subsection{Geometry-Weighted Contrastive Learning}

GeoCAM aligns sonar and optical embeddings using a geometry-weighted InfoNCE objective. Standard contrastive learning assigns a uniform weight to all positive pairs, but in opti-acoustic surveys the correspondence strength depends on the actual overlap between sonar and optical footprints. GeoCAM therefore uses overlap ratios as soft weights, providing stronger alignment pressure to pairs that share more seabed content.

For a batch of $B$ sonar tiles and their corresponding optical descriptors, we compute the similarity matrix
\begin{equation}
S_{ij} = \mathrm{sim}(z^s_i, z^o_j) / \tau,
\label{eq:similarity_matrix}
\end{equation}
where $\mathrm{sim}(\cdot,\cdot)$ is cosine similarity and $\tau$ is the temperature.

The geometry-weighted contrastive loss is defined as
\begin{equation}
\begin{aligned}
\mathcal{L}_{\mathrm{GW\text{-}InfoNCE}} = -\frac{1}{2B}\sum_{i=1}^{B} w_i
\Bigg(
&\log \frac{\exp(S_{ii})}{\sum_{k=1}^{B}\exp(S_{ik})} \\
&+ \log \frac{\exp(S_{ii})}{\sum_{k=1}^{B}\exp(S_{ki})}
\Bigg).
\end{aligned}
\label{eq:gw_infonce}
\end{equation}
where $w_i \in [0.4, 1.0]$ is the geographic overlap weight for the $i$-th 
matched sonar-optical pair, clamped to avoid suppressing low-overlap 
positives entirely. The two log terms correspond to the sonar$\rightarrow$optical 
and optical$\rightarrow$sonar directions respectively, and $S_{ij} = 
\cos(z^s_i, z^o_j)/\tau$.

\subsection{Cross-Modal Masked Reconstruction}

GeoCAM supports an optional cross-modal masked reconstruction (CMMR) branch to further regularize the learned representations. The goal of this auxiliary task is to reconstruct masked sonar patches using optical context, encouraging the shared embedding space to retain physically meaningful seabed information.

When CMMR is enabled, a portion of sonar patches is randomly masked in pixel space, while the remaining visible patches are encoded by the sonar backbone. The aggregated optical embedding provides cross-modal context to a lightweight decoder, which predicts the missing sonar content. We mask 40\% of the sonar patches, which is lower than typical MAE \cite{he2022mae} settings to preserve reliable global sonar features for contrastive alignment. In our final retrieval recipe, CMMR is turned off and sonar tiles are encoded without masking.

Let $x_p$ denote a masked sonar patch and $\hat{x}_p$ its prediction. The reconstruction loss is defined as
\begin{equation}
\mathcal{L}_{\mathrm{CMMR}} =
\frac{1}{|\mathcal{M}|}
\sum_{p \in \mathcal{M}}
\left\|
\hat{x}_p -
\frac{x_p - \mu_p}{\sqrt{\sigma_p^2 + \epsilon}}
\right\|_2^2,
\label{eq:cmmr_loss}
\end{equation}
where $\mathcal{M}$ is the set of masked patch indices and $\mu_p$, $\sigma_p^2$ are computed per patch. The decoder is discarded after pre-training and does not affect inference-time computation.

The overall training objective is
\begin{equation}
\mathcal{L}_{\mathrm{total}} =
\mathcal{L}_{\mathrm{GW-InfoNCE}} + \lambda \mathcal{L}_{\mathrm{CMMR}},
\label{eq:total_loss}
\end{equation}
where $\lambda$ controls the contribution of the auxiliary reconstruction term. Based on ablation results, we set $\lambda=0$ in the final retrieval-oriented recipe, as contrastive-only training yields the most stable and effective performance.

\section{Experiments}
\label{sec:experiments}

\subsection{Datasets}
\label{sec:datasets}

We evaluate GeoCAM on the BenthiCat side-scan sonar with co-registered camera (SSS-CAM) unlabeled pairing benchmark~\cite{rajani2025benthicat}. The benchmark comprises 2,080 SSS tiles geo-referenced against 35,947 matched optical 
images, together with correspondence metadata and survey geometry for cross-modal pairing. Following the released SSS-CAM protocol, we use only the unlabeled pairing assets and do not rely on semantic labels or segmentation masks in the experiments reported in this paper.

We split the data into 80\% training and 20\% validation subsets using a fixed random seed for reproducibility. During training, each sonar tile is associated with up to eight optical views. Geographic overlap priors are computed offline from the survey metadata and used as soft weights in the contrastive objective.

\subsection{Evaluation Metrics}
\label{sec:evaluation_metrics}

We report the following standard cross-modal retrieval and ranking metrics on the validation split: 
\begin{itemize}
    \item \textbf{Recall@$k$ (R@$k$)} is equal to 1 when the geographically matched paired item appears among the top-k retrieved gallery neighbors, and 0 otherwise, averaged over queries. We report $k\in\{1,5,10\}$.
    \item \textbf{Mean Reciprocal Rank (MRR)} is computed as the average of $1/\mathrm{rank}$ over queries, where $\mathrm{rank}$ is the 1-based position of the correct match in the retrieved list. We report directional MRR when needed.
    \item \textbf{Normalized Discounted Cumulative Gain (NDCG@$k$)} measures ranking quality with position discounting. We report NDCG@5 for the sonar$\rightarrow$optical direction when a directional score is required.
    \item \textbf{Alignment diagnostics} In addition to retrieval metrics, we log the mean positive and negative similarities and their difference (\texttt{pos\_minus\_neg}), which summarizes the separation between matched pairs and negatives in the learned embedding space.
\end{itemize}

Retrieval is evaluated in both sonar$\rightarrow$optical and optical$\rightarrow$sonar directions, and scalar values denote the mean of the two directions unless stated otherwise.

\subsection{Implementation Details and Computational Resources}
\label{sec:implementation_details}



GeoCAM is optimized end-to-end using AdamW~\cite{loshchilov2019adamw} with encoder and projection-head learning rates of $5\times10^{-5}$ and $5\times10^{-4}$, respectively, weight decay $0.05$, batch size $64$, 100 epochs, 10-epoch warm-up, cosine decay, gradient clipping (max-norm $1.0$), mixed-precision training, and temperature $\tau=0.15$.
The best checkpoint is selected by validation loss, while the cross-modal masked reconstruction branch and hard negative mining are disabled in the final recipe.
Sonar tiles are resized to $256\times256$, and optical views are encoded by a frozen DINOv2 ViT-S/14~\cite{oquab2024dinov2} and aggregated with SAA before projection into the shared embedding space.
All experiments are conducted on three NVIDIA TITAN RTX 24\,GB GPUs, and final results are reported as mean$\,\pm\,$std over multiple random seeds.

\section{Results and Discussion}
\label{sec:results}

We evaluate GeoCAM in the unlabeled cross-modal retrieval setting, where the task is to retrieve the correct optical views given a sonar query. 

\subsection{Retrieval Performance and Stability}

GeoCAM is highly stable across independent training runs. Table~\ref{tab:best_three_runs} summarizes the best checkpoints selected by validation loss, and the three runs achieve consistent results, with Recall@1 between 0.28 and 0.31, Recall@10 ranging from 0.89 to 0.92, and closely matched MRR and NDCG@5 values. This consistency indicates that GeoCAM learns a robust alignment space rather than depending on a favorable initialization or an isolated training trajectory. The t-SNE projection in Fig.~\ref{fig:architecture} (panel E) further confirms that matched sonar and optical embeddings cluster together in the shared space, while 
non-matching pairs remain well separated.
\begin{table}[t]
\centering
\small
\caption{Performance of GeoCAM across three seeds (mean $\pm$ std in final row)}
\label{tab:best_three_runs}
\resizebox{\columnwidth}{!}{%
\begin{tabular}{lccccc}
\hline
Runs & R@1 (\%) & R@5 (\%) & R@10 (\%) & MRR\textsuperscript{*} (\%) & NDCG@5\textsuperscript{*} (\%) \\
\hline
GeoCAM\textsuperscript{1} & 28.9 & 78.1 & 91.5 & 50.3 & 55.6 \\
GeoCAM\textsuperscript{2} & 31.1 & 76.1 & 88.6 & 50.0 & 54.9 \\
GeoCAM\textsuperscript{3} & 28.2 & 77.5 & 92.2 & 49.5 & 55.4 \\
\hline
Mean $\pm$ std & 29.4$\pm$1.5 & 77.2$\pm$1.1 & 90.8$\pm$1.9 & 49.9$\pm$0.4 & 55.3$\pm$0.4 \\
\hline
\end{tabular}%
}\\
{\raggedright \footnotesize \;\; 1: seed42 - 2: seed52 - 3: seed62 - *: sonar$\rightarrow$optical \par}
\end{table}

\subsection{Ablation Study}

The ablation study in Table~\ref{tab:ablation_study} clarifies how GeoCAM reaches this behavior. As no prior cross-modal method has been evaluated on this newly released benchmark, we use the mean-pooling baseline as a reference point; it already performs reasonably well, confirming that BenthiCat SSS-CAM contains exploitable unlabeled correspondences. However, SAA alone does not improve retrieval, which suggests that set-level view selection is not sufficient without geometric supervision. In contrast, Geometry-Weighted InfoNCE improves alignment and ranking quality, supporting the central hypothesis of GeoCAM: the continuous footprint overlap ratio is a meaningful supervision signal and should be used directly in contrastive learning. Hard negative mining slightly reduces retrieval performance, while the CMMR branch substantially degrades performance and is therefore not retained in the final recipe.

\begin{table*}[t]
\centering
\caption{Ablation study for GeoCAM under a controlled short-budget training protocol (20 epochs)\textsuperscript{\dag}.
All variants share the same hyperparameters and are compared under identical conditions.}
\label{tab:ablation_study}
\resizebox{\textwidth}{!}{
\begin{tabular}{lccccccccccc}
\hline
Method Variant & SAA & GeoW InfoNCE & CMMR & HardNeg & Temp & R@1 (\%) & R@5 (\%) & R@10 (\%) & MRR (\%) & NDCG@5 (\%) & Pos$-$Neg (\%) \\
\hline
MeanPool Contrastive Baseline        & \texttimes & \texttimes  & \texttimes & \texttimes & 0.15 & 19.5 & 59.0 & 79.0 & 37.0 & 38.0 & 60.0 \\
SAA Only                             & \checkmark & \texttimes  & \texttimes & \texttimes & 0.15 & 17.5 & 56.0 & 76.0 & 34.5 & 34.5 & 63.0 \\
SAA + Geo-Weighted InfoNCE           & \checkmark & \checkmark  & \texttimes & \texttimes & 0.15 & 20.0 & 59.0 & 78.5 & 36.5 & 37.5 & 61.5 \\
SAA + Geo-Weighted InfoNCE + CMMR    & \checkmark & \checkmark  & \checkmark & \texttimes & 0.15 &  4.0 & 12.0 & 20.0 &  9.0 &  7.0 & 38.0 \\
SAA + Geo-Weighted InfoNCE + HardNeg & \checkmark & \checkmark  & \texttimes & \checkmark & 0.15 & 18.5 & 56.0 & 76.0 & 34.5 & 34.5 & 60.5 \\
\hline
GeoCAM (Full Recipe, 100ep)          & \checkmark & \checkmark  & \texttimes & \texttimes & 0.15 & 28.9 & 78.1 & 91.5 & 50.3 & 55.6 & 73.3 \\
\hline
\end{tabular}
}
\\[2pt]
{\raggedright \footnotesize \;\; \textsuperscript{\dag}Full model uses the complete 100-epoch recipe (Sec.~3.3); ablation variants here are trained for 20 epochs to enable efficient comparison across configurations. \par}
\end{table*}
\subsection{Qualitative Retrieval Analysis}

\begin{figure*}[t]
\centering
\includegraphics[width=0.92\textwidth]{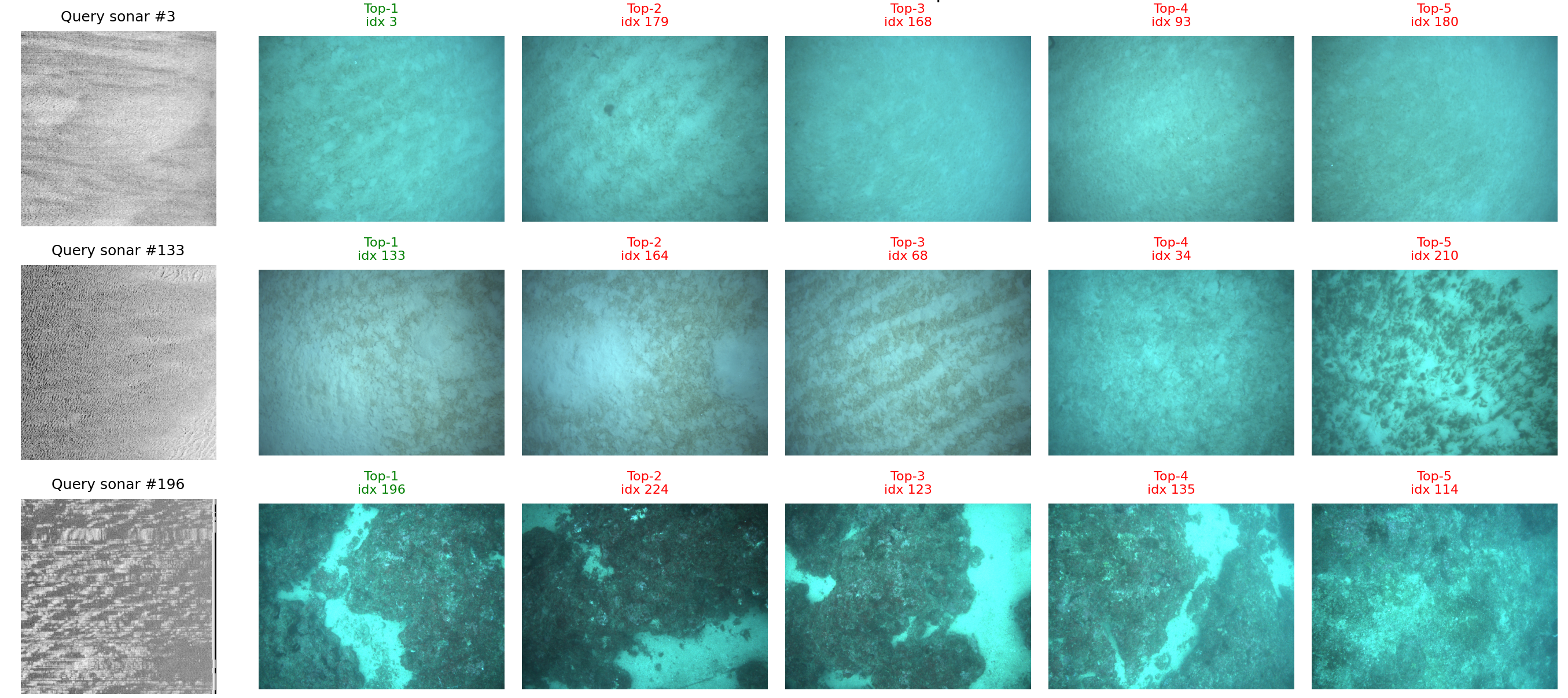}
\caption{Three representative cross-modal retrieval examples produced by GeoCAM.}
\label{fig:retrieval_examples}
\end{figure*}

Figure~\ref{fig:retrieval_examples} provides qualitative evidence consistent with the quantitative results. For the three sonar queries, the retrieved optical views are visually consistent with the sonar texture and seabed structure. Smooth sandy or rippled regions retrieve homogeneous optical scenes, while more complex sonar patterns retrieve images with denser substrate appearance. These examples suggest that GeoCAM learns seabed-relevant cross-modal correspondence rather than superficial visual similarity.

\subsection{Discussion}

Two observations emerge clearly from these results. First, geographic footprint overlap provides a meaningful form of soft supervision, allowing GeoCAM to treat positive correspondences as a continuous signal rather than a binary assignment. Second, the one-to-many structure of the dataset must be modeled explicitly, since each sonar tile is associated with multiple optical frames of varying quality. Together, these two design choices explain the improvement over mean-pooling baselines and motivate the use of overlap-aware set aggregation in opti-acoustic retrieval.

The CMMR branch does not improve retrieval in the current setting. A plausible explanation is that reconstruction encourages low-level sonar appearance preservation, which can conflict with discriminative cross-modal alignment. It may also introduce optimization instability under limited correspondence data.

\subsection{Limitations and Future Work}

This study focuses exclusively on unlabeled cross-modal alignment and retrieval and does not yet evaluate GeoCAM as a pretraining strategy for downstream supervised tasks. Although BenthiCat provides segmentation masks and semantic annotations, the current experiments operate only on the SSS-CAM correspondence subset without exploiting class labels. Future work will integrate the labeled split to assess whether GeoCAM improves label-efficient benthic habitat classification and semantic segmentation. In addition, extending the framework toward sonar-to-sonar matching and cross-sector retrieval could further strengthen its applicability to large-scale seafloor mapping, where optical imagery may not be available at inference time.
\section{Conclusions}
\label{sec:conclusion}

GeoCAM demonstrates that the geometric metadata inherent to standard opti-acoustic surveys — specifically, GPS-derived footprint overlap ratios constitutes an effective and zero-cost supervision signal for cross-modal self-supervised learning.
By combining set-level optical aggregation with geometry-weighted contrastive supervision, the framework aligns sonar tiles and optical views in a shared embedding space without any semantic annotations, establishing a reproducible pre-training baseline on BenthiCat SSS-CAM.

Experiments confirm stable retrieval performance across independent runs and consistent improvement over ablation variants when geometry-aware supervision is applied.
Set-aware modeling proves necessary to handle the one-to-many optical correspondence structure, while reconstruction-based auxiliary training degrades performance in the retrieval-focused setting.
Future work will extend GeoCAM to supervised downstream tasks, benthic segmentation, and cross-sector sonar matching to assess its value as a pre-training strategy for label-efficient marine habitat understanding.

{\scriptsize
\bibliographystyle{ieee}
\bibliography{egbib}
}

\end{document}